\documentclass[sigconf]{acmart}
\usepackage{latexsym}
\usepackage{amsmath}
\usepackage{graphicx}
\usepackage{booktabs}
\usepackage{multirow}
\usepackage{balance}

\usepackage{fancyvrb}
\usepackage{todonotes}
\usepackage{microtype}

\newcommand\BibTeX{B\textsc{ib}\TeX}
\usepackage{tikz}
\DeclareRobustCommand*\circled[1]{\tikz[baseline=(char.base)]{
            \node[shape=circle,draw,inner sep=2pt] (char) {#1};}}
\AtBeginDocument{%
  \providecommand\BibTeX{{%
    \normalfont B\kern-0.5em{\scshape i\kern-0.25em b}\kern-0.8em\TeX}}}

\copyrightyear{2022}
\acmYear{2022} 
\setcopyright{rightsretained}
   
\acmConference[CIKM '22]{Proceedings of the 31st ACM International Conference on Information and Knowledge Management}{October 17--21, 2022}{Atlanta, GA, USA}
\acmBooktitle{Proceedings of the 31st ACM International Conference on Information and Knowledge Management (CIKM '22), October 17--21, 2022, Atlanta, GA, USA}
\acmPrice{15.00}
\acmDOI{10.1145/3511808.3557196} 
\acmISBN{978-1-4503-9236-5/22/10}

\usepackage{etoolbox}
\makeatletter
\patchcmd{\maketitle}{\@copyrightpermission}{
   \begin{minipage}{0.3\columnwidth}
     \href{https://creativecommons.org/licenses/by/4.0/}{\includegraphics[width=0.90\textwidth]{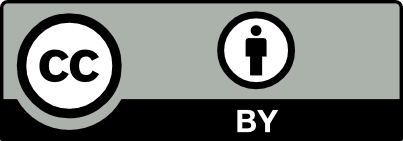}}
   \end{minipage}\hfill
   \begin{minipage}{0.7\columnwidth}
     \href{https://creativecommons.org/licenses/by/4.0/}{This work is licensed under a Creative Commons Attribution International 4.0 License.}
   \end{minipage}

   \vspace{5pt}
}{}{}

\makeatother

\acmSubmissionID{demo123}

\begin{document}
\sloppy
\title{POTATO: exPlainable infOrmation exTrAcTion framewOrk}

\author{\'Ad\'am Kov\'acs}
\email{adam.kovacs@tuwien.ac.at}
\affiliation{%
  \institution{TU Wien}
  \streetaddress{Favoritenstra{\ss}e 9-11.}
  \city{Vienna}
  \country{Austria}
  \postcode{1040}
}
\affiliation{%
\institution{Budapest University of Technology and
	Economics}
\streetaddress{Műegyetem rkp. 3.}
\city{Budapest}
\country{Hungary}
\postcode{H-1111}
}

\author{Kinga G\'emes}
\email{kinga.gemes@tuwien.ac.at}
\affiliation{%
	\institution{TU Wien}
	\streetaddress{Favoritenstra{\ss}e 9-11.}
	\city{Vienna}
	\country{Austria}
	\postcode{1040}
}
\affiliation{%
\institution{Budapest University of Technology and
	Economics}
\streetaddress{Műegyetem rkp. 3.}
\city{Budapest}
\country{Hungary}
\postcode{H-1111}
}

\author{Eszter Ikl\'odi}
\email{eszter.iklodi@tuwien.ac.at}
\affiliation{%
  \institution{TU Wien}
  \streetaddress{Favoritenstra{\ss}e 9-11.}
  \city{Vienna}
  \country{Austria}
  \postcode{1040}
}

\author{G\'abor Recski}
\email{gabor.recski@tuwien.ac.at}
\affiliation{%
  \institution{TU Wien}
  \streetaddress{Favoritenstra{\ss}e 9-11.}
  \city{Vienna}
  \country{Austria}
  \postcode{1040}
}

% %%
% %% By default, the full list of authors will be used in the page
% %% headers. Often, this list is too long, and will overlap
% %% other information printed in the page headers. This command allows
% %% the author to define a more concise list
% %% of authors' names for this purpose.
\renewcommand{\shortauthors}{Ádám Kovács, Kinga Gémes, Eszter Iklódi, \& Gábor Recski}
%% If needed use a foot or author note to identify equal contribution

\begin{abstract}
  We present \texttt{POTATO}, a task- and language-independent framework for
human-in-the-loop (HITL) learning of rule-based text classifiers using
graph-based features.  POTATO handles any type of directed graph and
supports parsing text into Abstract Meaning Representations (AMR),
Universal Dependencies (UD), and 4lang semantic graphs.  A web-based
user interface allows users to build rule systems from graph patterns,
provides real-time evaluation based on ground truth data, and suggests
rules by ranking graph features using interpretable machine learning
models. Users can also provide patterns over graphs using regular
expressions, and POTATO can recommend refinements of such rules. POTATO is
applied in projects across domains and languages, including classification
tasks on German legal text and English social media data.  All components
of our system are written in Python, can be installed via pip, and are
released under an MIT License on GitHub.
\end{abstract}





\maketitle

\section{Introduction}
Recent natural language processing (NLP) solutions achieving state-of-the-art
results on public benchmarks rely on deep learning models with millions of
parameters. Since such models require large amounts of training data, offer
little or no explainability of their decisions, and pose a risk of learning
unintended bias, their applicability in real-world scenarios is limited.
Rule-based systems may provide accurate and transparent solutions, but can be
difficult to build and maintain.  POTATO is a framework that supports the
semi-automatic creation of rule-based text classifiers. Firstly, rules can be
formulated as patterns over graphs representing the syntax and/or semantics of
the input text. Secondly, ground truth labels are used to suggest rules to the
users which they may choose to accept with or without modifications. POTATO can
also recommend refinements of underspecified graph patterns, allowing the user
to guide the rule learning process. In addition to a framework for the
automatic extraction and ranking of graph-based features, POTATO also provides
an intuitive, high-level user interface that allows users without technical
expertise to construct graph-based rule systems using automatic suggestions and
to evaluate rules against ground truth data in real time.  We also provide a
REST API to our backend for deploying rule-based solutions built in POTATO.
All components of our system are released under an MIT license and are
available from GitHub\footnote{\url{https://github.com/adaamko/POTATO}} and via
\texttt{pip}, by installing the
\texttt{xpotato}\footnote{\url{https://pypi.org/project/xpotato/}} package. A
short video demonstration is available on
Youtube\footnote{\url{https://youtu.be/79Q7UnLsMFc}}.
The rest of this paper is structured as follows. Section~\ref{sec:rel} reviews
recent systems for rule learning and HITL learning. Section~\ref{sec:method}
presents our method for generating and ranking graph-based features from
labeled datasets. Section~\ref{sec:ui} presents our web-based user
interface for HITL learning, inspection, editing, and evaluation of rule
systems. An overview of the system architecture and workflow is provided in
Section~\ref{sec:system}. In Section~\ref{sec:app} we showcase recent
applications of our system across domains and languages. 

\section{Related work}
\label{sec:rel}

Our work presents a semi-automatic human-in-the-loop (HITL) method for the
learning of rule systems for text classification tasks using patterns over
graph representations of natural language input. The graphs we use for
representing the syntax and semantics of text include Universal Dependencies
\cite[UD, ][]{Nivre:2018s}, Abstract Meaning Representations \cite[AMR,
][]{Banarescu:2013}, and 4lang concept graphs \citep{Kornai:2019}.  Recent approaches
to rule learning for text classification include the learning of first-order
logic formulae over semantic representations using neural networks
\cite{Sen:2020,Evans:2018} and integer programming \cite{Dash:2018}.  For
suggesting rules to users based on ground truth labels we rely on interpretable
machine learning models such as decision trees.  Other rule-based approaches to text classification may
involve the use of textual patterns \cite{Lertvittayakumjorn:2021}, semantic
structures \cite{Sen:2019}, or a combination of regular expressions and graph
patterns \cite{Valenzuela:2020}.  Hybrid approaches to text classification
include the incorporation of lexical features into DL architectures
\cite{Koufakou:2020,Pamungkas:2019} and voting between rule-based and ML
systems \cite{Razavi:2010,Gemes:2021b}.  The systems most similar to POTATO are
the HEIDL \citep{Sen:2019} and GrASP
\citep{Lertvittayakumjorn:2021,Shnarch:2017} libraries, both of which support
pattern-based text classification with automatic suggestions.  Our tool differs
from these in its use of syntactic and semantic graphs to represent text input,
its ability to suggest refinements of rules specified by the user, and
functionalities for working with data that has little or no ground truth
annotation available. We shall describe all features of our system in detail in
Sections~\ref{sec:method}~and~\ref{sec:ui}.

\section{Method} 
\label{sec:method}

In this section we present the core functionalities of POTATO. As a running
example throughout the section we shall use the relation extraction task defined
for the 2010 Semeval task \textit{Multi-Way Classification
of Semantic Relations Between Pairs of Nominals} \citep{Hendrickx:2010},
In this task, sentences containing two entities identified by
the labels \texttt{Entity1} and \texttt{Entity2} must be classified based on 
the semantic relationship that holds between those entities. In our example we
focus on the most frequent label in the dataset, \texttt{Entity-Destination (ED)},
which is defined as follows: \textit{``An entity is moving towards a
destination. Example: the \underline{boy} went to \underline{bed}"}
\cite[p.34]{Hendrickx:2010}. The training data for this task
contains 8000 sentences, 844 of which are labeled as examples of
\texttt{ED}. We used 80\% of this dataset for developing our
rule system and the remaining 20\% for validation.
The rules we construct using POTATO are patterns over \texttt{4lang} graph representations of the
sentences, we construct these graphs using the implementation of
\texttt{text\_to\_4lang} \cite{Recski:2018} in the
\texttt{tuw\_nlp}\footnote{\url{https://pypi.org/project/tuw-nlp/}} library.

For each class label, POTATO enables the construction of a list of rules that are used to
label sentences as belonging to that class if and only if at least one rule
applies to the graph representation of the sentence. A single rule may be a
pattern over a graph or a conjunction of multiple patterns, some of which may
be negated. A single pattern is a graph that matches the input graph if and
only if the pattern graph is a subgraph of the input graph. Graph
nodes and edges may have string labels, and pattern graphs may contain regular
expressions (regexes) as both edge and node labels. If a pattern graph contains
regex labels, then they match an input graph if and only if the pattern graph
is contained in the input graph such that the regex labels of the pattern graph
match the string labels of corresponding nodes and edges in the input graph.
Since a rule may be a conjunction of patterns, each of which may be negated,
and a rule system is a disjunction over rules, a complete rule system can be
considered the disjunctive normal form (DNF) of a single boolean formula
whose predicates are the individual graph patterns.

POTATO provides rule suggestions by training interpretable machine learning
models using graph features. The features extracted from each graph are
its connected subgraphs of at most $n$ edges. In most applications we set $n$ to 2.
Graphs consisting of a single node are included for any value of $n$, since
they are connected graphs with 0 edges.
A training dataset of labeled texts that are each represented by graphs can then
be used to train interpretable ML models and rank subgraphs based on their
feature importance. The core assumption of our method is that subgraphs with
high feature importance are the ones that should be suggested to the user
constructing a rule system. Currently POTATO trains decision trees as implemented by the
library \texttt{scikit-\-learn} \cite{Pedregosa:2011} and ranks subgraphs based on
their Gini coefficients. The system also includes the option of
ranking subgraphs without training any ML model, by counting the number of
positive and negative examples in the training data that contain each subgraph
as a feature and calculating $TP - FP$, i.e. the difference between the number
of true positive and false positive decisions one would make by classifying
input sentences based on the presence of this pattern only. Top-ranked
subgraphs are presented to the user, who can choose to incorporate them in
the rule system, with or without manual modifications.

Using the Semeval relation extraction dataset for training, the top suggestion
without manual modifications would be a graph containing the single edge
\textit{into~$\xrightarrow{2}$~ENTITY2} (edges labeled 1 and 2 in 4lang graphs
mark the first and second arguments of binary relations).  This rule in itself
achieves 76.2\% precision and 62.8\% recall on the training data, retrieving
407 true positive and 127 false negative samples.  A human expert might extend
this pattern
by constructing the graph
\textit{ENTITY2~$\xleftarrow{2}$~into~$\xrightarrow{1}$~(.*)~$\xrightarrow{2}$~ENTITY1},
which requires that \textit{ENTITY1} is the second argument of the concept that is the
first argument of the \textit{into} relation.
POTATO uses the PENMAN notation \citep{Goodman:2020} to represent graphs on its
frontend, the original graph would be displayed as {\small
\texttt{(u\_1~/~into~:2~(u\_2~/~entity2))}} and the modified graph would be
entered by the user as {\small \texttt{(u\_1~/~into~:2 (u\_2~/~entity2)~:1
(u\_3~/~.*~:2 (u\_4~/~entity1)))}}. This modified graph now achieves 85.3\%
precision and 41.0\% recall for the \texttt{ED} label,
by matching 266 true positives (such as the pair of entities in the sentence
\textit{We have dumped the \textbf{spam} into the \textbf{junk folder}}) and 46
false positives (such as \textit{The \textbf{lungs} are divided into
\textbf{lobes}}). The ranking of subgraphs by feature importance can also be
used to suggest refinements of a subgraph containing regular expressions.  If
the user creates or modifies a rule so that some label contains a regular
expressions and asks the system to \textit{refine} this rule, POTATO will
evaluate matching subgraphs and replace the regex with a disjunction of all
string labels with which the pattern would achieve over 90\% precision (and
non-zero recall) on the training dataset.
% Refining the regex node of the
% previous rule yields the following pattern:
% \begin{Verbatim}[fontsize=\scriptsize]
% (u_1 / into
%   :2 (u_2 / entity2)
%   :1 (u_3 / misplace|invest|drop|import|transport|
%             fetch|pack|insert|pour|implant|remove|
%             leak|add|dump|release|stuff
%        :2 (u_4 / entity1)))
% \end{Verbatim}
Refining the regex node of the previous rule would yield a pattern that would match 138
true positives and only 1 false positive (\textit{A US \textbf{aircraft} was dropped
into a difficult \textbf{landing} in Mali}), achieving 99.3\% precision on the training dataset. On
the validation data the same pattern achieves 95.6\% precision and 21.9\%
recall\footnote{the list of concepts constructed by POTATO for this rule consists of
these 16 words: \textit{ misplace, invest, drop, import, transport, fetch, pack, insert,
pour, implant, remove, leak, add, dump, release, stuff }}. 
%\todo{remaining problems, "conclusion"} False positives include The EU inserted
%<e1>experts</e1> into the national <e2>defence</e2>

\section{User interface} \label{sec:ui}

\begin{figure}[htbp]
  \begin{center}
      \includegraphics[width=0.48\textwidth]{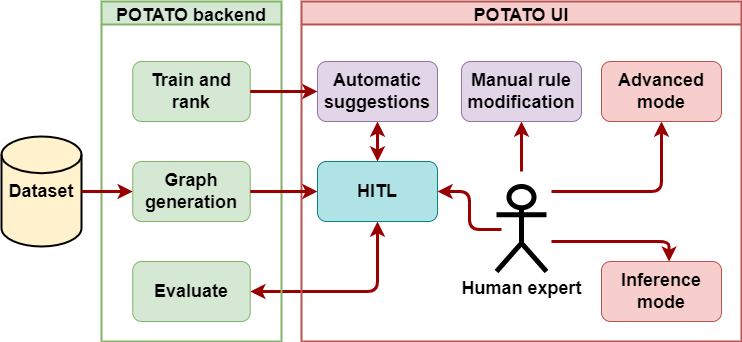}
  \end{center}
  \caption{System workflow of POTATO}
  \label{fig:potato_workflow}
\end{figure}

\begin{figure*}[htbp]
  \begin{center}
      \includegraphics[width=\textwidth]{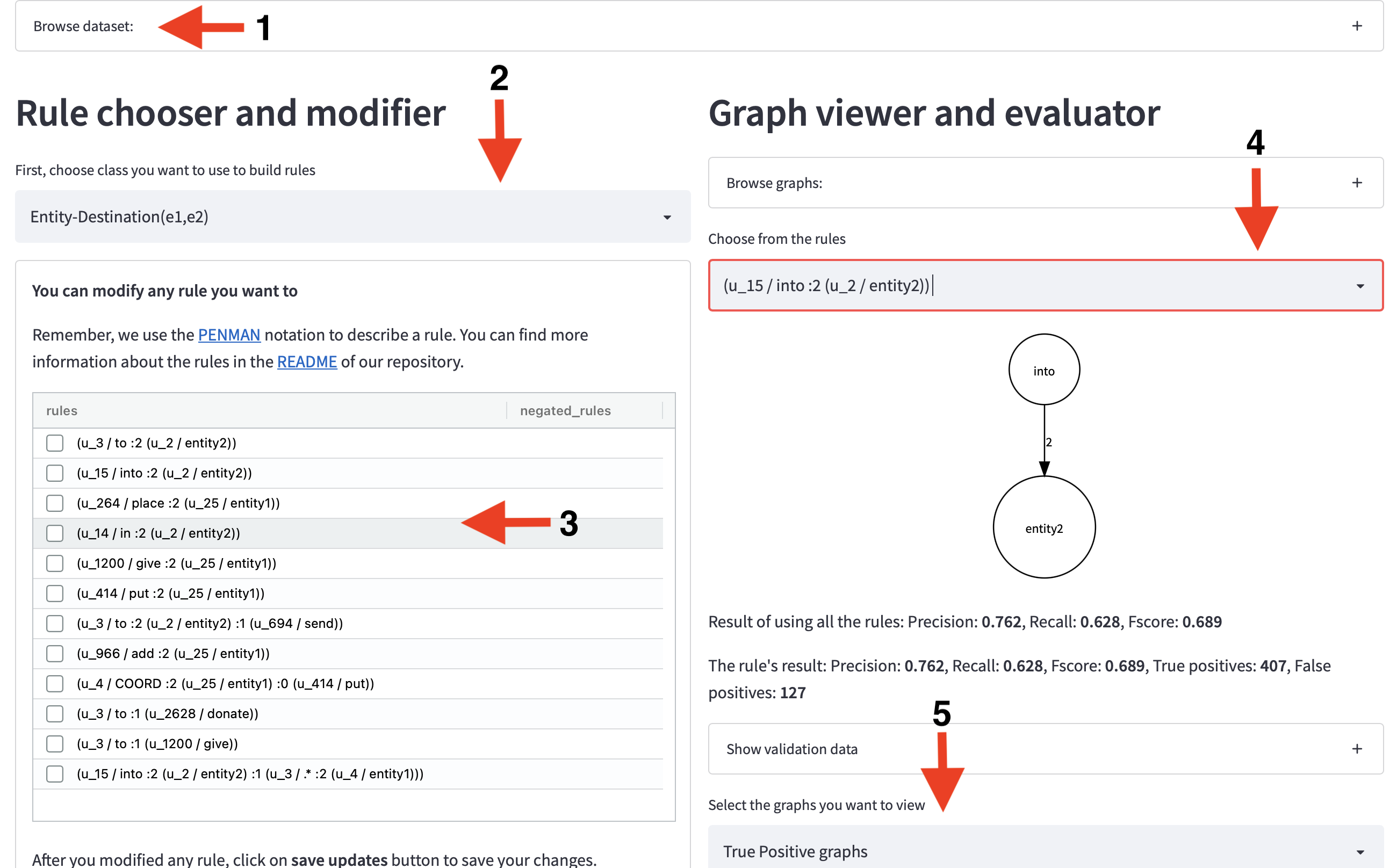}
  \end{center}
  \caption{The main page of POTATO allows the user to \circled{1} browse the dataset and view the processed graphs \circled{2} choose the
  class you want to build rule-based systems on \circled{3} modify, delete, add new rules and get suggestions \circled{4} view the results of the selected rules
  \circled{5} view example predictions for each rule}
  \label{fig:potato_UI}
\end{figure*}

Figure~\ref{fig:potato_workflow} illustrates the human-in-the-loop (HITL)
workflow enabled by our application. The prerequisite for launching the HITL
user interface is to load a dataset as a set of labeled or
unlabeled graphs. Any directed graph can be loaded, we provide interfaces for
UD parsers in \texttt{stanza}, two AMR parsers for English and German, and our own
language-agnostic 4lang parser. Suggesting and evaluating rules requires
ground truth labels, if these are not available, the UI can be launched in
\textit{advanced} mode for bootstrapping labels using rules. Once a dataset is
loaded, the HITL frontend can be started and the user is presented with the
interface shown in Figure~\ref{fig:potato_UI}, built using the
\texttt{streamlit} library\footnote{\url{https://streamlit.io/}}.

The dataset
browser \circled{1} allows the user to view the text, graph, and label for all
rows of the dataset. The viewer renders graphs using the \texttt{graphviz}
library \cite{Gansner:2000}, and also provides the PENMAN notation that can be
copied by the user for quick editing of rules.  Users can choose the class to
work on \circled{2}, the list of rules constructed for each class are
maintained in a list \circled{3}. Rules can be viewed and evaluated on the
training and validation datasets \circled{4}, and users can analyze correct and
incorrect predictions of each rule by choosing to view true positive, false
positive, or false negative examples \circled{5}.
Functions not visible on Figure~\ref{fig:potato_UI} due to lack of space include
the button
\textit{suggest new rules}, which returns a list of suggested graphs together with
their performance on the training data, allowing the user to select those that
should be added to the rule list.
% this interface is shown in Figure~\ref{fig:potato_suggestions}. 
For rules containing regular expressions,
the \textit{Refine} button will replace regular expressions with a disjunction
of high-precision labels, as described in Section~\ref{sec:method}. Rules are automatically
saved and can be reloaded. Rules then also can be used for inference (from the UI, from the command line, or via a REST API).

% Rules for
% each class are automatically saved to disk in JSON format, this file can be
% loaded for further editing or for inference. Using a rule system for predicting
% labels is possible both on the UI and by launching a command-line tool
% implementing a service with a REST API. 

%\begin{figure*}[htbp]
  %\begin{center}
      %%\includegraphics[width=0.49\textwidth,trim={1cm 1cm 0.5cm 0.2cm}]{../files/potato_suggestions.PNG}
      %\includegraphics[width=\textwidth]{../files/potato_suggestions_clipped.png}
  %\end{center}
  %\caption{Patterns suggested by POTATO, ranked by precision}
  %\label{fig:potato_suggestions}
%\end{figure*}

\section{System Architecture}
\label{sec:system}

% \begin{figure}[htbp]
%   \begin{center}
%       \includegraphics[width=0.49\textwidth,trim={1cm 1cm 0.5cm 0.2cm}]{../files/potato_architecture.png}
%   \end{center}
%   \caption{The architecture of POTATO}
%   \label{fig:potato_architecture}
% \end{figure}

%Figure~\ref{fig:potato_architecture} shows the high-level architecture of the POTATO system. 
All software is released via two python packages
(\texttt{xpotato}, \texttt{tuw-nlp}), both installable via pip. The frontend of 
the system is included as a separate module of the \texttt{xpotato} library.
In this section we describe each component of our system.

\paragraph{tuw-nlp} The tuw-nlp package contains a suite of generic NLP tools
and includes the \texttt{graph} module for building and manipulating linguistic graphs.
The module uses networkx\footnote{\url{https://networkx.org/}} for its 
graph data structures,
and implements
the graph operations used by POTATO's backend, including the
generation of subgraphs and the pattern matching on graphs for the evaluation
and ranking of graph patterns. The pattern matcher class customizes and wraps
the \texttt{DiGraphMatcher} class from
\texttt{networkx.algorithms.isomorphism}, which implements the vf2 algorithm
\cite{Cordella:2001}.
The \texttt{tuw\_nlp.graph} module also provides interfaces for working with
AMR graphs (English or German), UD graphs, and 4lang semantic graphs.
For English AMR-parsing, the module interfaces with
a pretrained Transformer-based AMR parser~\cite{Raffel:2020} via the
\texttt{amrlib}\footnote{\url{https://amrlib.readthedocs.io/en/latest/}} library,
German AMRs are constructed using a multilingual,
transition-based system \cite{Damonte-Cohen:2018} implemented by the \texttt{amr-eager-multilingual}\footnote{\url{https://github.com/mdtux89/amr-eager-multilingual}} library.
For Universal Dependency (UD) parsing, \texttt{tuw-nlp} relies on the
\texttt{stanza} package \cite{Qi:2020}, and 4lang semantic graphs are built
from UD trees using a reimplementation
of the \texttt{text\_to\_4lang} tool \citep{Recski:2018}.

\paragraph{xpotato}
The backend of POTATO in the main module of the \texttt{xpotato} package
implements core functionalities by interfacing with the \texttt{tuw\_nlp.graph}
module described above, the \texttt{scikit\-learn} library \cite{Pedregosa:2011}
for training and inspecting decision trees and the \texttt{scikit-\-criteria}
package \citep{Cabral:2016} for feature ranking. The module also implements the
algorithm behind the \textit{refine} mechanism (see Section~\ref{sec:method})
for suggesting refinements of graph patterns containing regular expressions.
The frontend of POTATO is a web application implemented using \texttt{streamlit}
that exposes all features of the main \texttt{xpotato} module and enables the worflow
described in Section~\ref{sec:ui}. 
% The graph viewer component uses the 
% \texttt{graphviz} library \cite{Gansner:2000} to render graphs into the
% webpage from the dot format. The data browser and rule browser components
% are implemented using
% \texttt{ag-grid}\footnote{\url{https://www.ag-grid.com/}}.
The web application can be started in three modes. The \textit{simple} mode corresponds
to the standard scenario when sufficient annotated training data
is available for the rule suggestion mechanism. 
% The \textit{advanced} mode
% contains all features of the simple mode and an additional frontend module
% for annotating unlabeled data, either manually or using graph patterns.
The \textit{advanced} mode contains an additional frontend module for annotating unlabeled
data supported by using graph patterns.
Launching the application in \textit{inference} mode loads a
saved list of rules and allows the user to apply the
rule set to raw text. Inference can also be run as a REST API using a command-line
tool, the service uses the FastAPI\footnote{\url{https://fastapi.tiangolo.com/}}
library.

\section{Applications}
\label{sec:app}

POTATO is language- and domain-agnostic. Besides the English relation
extraction task it has been used for
the classification of German legal documents and for offensive text detection
in English and German social media. In this section we describe these applications briefly.
For the task of relation extraction we have developed a simple rule-based
solution not only for the Semeval dataset introduced in
Section~\ref{sec:method}, but also a 
larger corpus for medical relation extraction called
CrowdTruth\footnote{\url{github.com/CrowdTruth/Medical-Relation-Extraction}}
\cite{Dumitrache:2017}. This benchmark contains two relations, \textit{treat}
and \textit{cause}. For the \textit{treat} relation, state-of-the-art deep learning
systems achieve F1-scores up to 0.9 \cite{Cenikj:2021}.
For a simple rule-based approach we used POTATO to build patterns
over UD graphs, and using only 12 rules we achieved precision and recall
values of 0.91 and 0.32, respectively, an example of a rudimentary but
high-precision and transparent solution. The rule system is available on
GitHub\footnote{\url{https://github.com/adaamko/POTATO/tree/main/features/crowdtruth}}.
%Recently, POTATO has also been applied to relation extraction from software
%documentation, as part of a system for deriving semantic validation rules from
%industrial standards specifications \cite{Bareedu:2022}. 

We also used POTATO to develop rule systems for the
2019-2021 HASOC datasets on offensive text detection,
\cite{Mandl:2019,Mandl:2020,Mandl:2021}. The rules for English were part of our
submission to the 2021 HASOC shared task \cite{Gemes:2021b}, where they were used to
increase the recall and F1 score of our top-performing deep learning models in an
ensemble solution. On the German HASOC dataset we achieve precision and recall values of
0.92 and 0.28 with just 8 graph patterns, these results are currently under review, the
rule-based solutions are available on
GitHub\footnote{\url{https://github.com/adaamko/POTATO/tree/main/features/hasoc}}.

Finally, POTATO is applied in the
BRISE\footnote{\url{https://smartcity.wien.gv.at/en/brise/}} project for
digitalizing the building permit process of the City of Vienna and involves
extracting formal rules from text documents of the city's Zoning Plan. A
crucial first step of the implemented pipeline \cite{Recski:2021} is the
multi-label classification of sentences based on the formal attributes of the
domain that they mention. For example, the sentence \textit{Flachd\"acher bis
zu einer Dachneigung von f\"unf Grad sind (\ldots) zu begr\"unen.} `Flat roofs with a pitch not
exceeding 5 degrees must be greened (\ldots)' must
be classified as mentioning three attributes, \texttt{BegruenungDach} `greening
of roofs', \texttt{Dachart} `roof type', and \texttt{DachneigungMax} `maximum
roof pitch'. While the total number of attributes is close to a hundred, the 20
most frequent ones cover nearly 75\% of all labels and the top 10 cover more
than 50\%. The latest rule-based solution to this task, available publicly on
GitHub\footnote{\url{https://github.com/recski/brise-plandok}} and to be published
in a forthcoming paper, has been implemented using \texttt{text\_to\_4lang}
and POTATO and on the 10 most frequent attributes achieves class-level
precision values above 0.8 and up to
1.0, class-level recall values between 0.5 and 0.95, and overall precision and
recall scores of 0.9 and 0.45, respectively. Early experiments show that
similar results can be obtained using standard machine learning algorithms,
but the rule system offers a fully transparent, configurable, and auditable
solution.

\begin{acks}
Work partially supported by BRISE-Vienna (UIA04-081), a European Union Urban Innovative Actions project.
\end{acks}

\bibliographystyle{ACM-Reference-Format}
\balance
\bibliography{tuw_nlp}

%%
%% If your work has an appendix, this is the place to put it

\end{document}